\documentclass{article}

\usepackage[preprint]{neurips_2026}

\usepackage[utf8]{inputenc}
\usepackage[T1]{fontenc}
\usepackage{hyperref}
\usepackage{url}
\usepackage{booktabs}
\usepackage{amsfonts}
\usepackage{amsmath}
\usepackage{amssymb}
\usepackage{bm}
\usepackage{nicefrac}
\usepackage{microtype}
\usepackage{xcolor}
\usepackage{graphicx}
\usepackage{multirow}

\usepackage{longtable}
\usepackage{array}
\usepackage{xurl}
\usepackage{ragged2e}
\usepackage{subcaption}
\usepackage{wrapfig}
\usepackage{float}

\hypersetup{hidelinks}
\usepackage{hyperref}
\hypersetup{
    colorlinks=true,
    urlcolor=blue,
    linkcolor=blue,
    citecolor=blue
}

\newcolumntype{S}{>{\RaggedRight\arraybackslash\scriptsize}p{0.31\linewidth}}
\newcolumntype{T}{>{\RaggedRight\arraybackslash\scriptsize}p{0.48\linewidth}}
\newcommand{\scene}[1]{\path{#1}}

\usepackage{wrapfig} %

\newcommand{\eg}{e.g.}

\title{NAMVIS: Next-Scale Autoregressive Multi-View Image Synthesis}

\author{%
  \textbf{Ramil Khafizov}$^{1}$ \quad
  \textbf{Ilya Statsenko}$^{2}$ \quad
  \textbf{Ruslan Rakhimov}$^{2}$ \\
  \textbf{Artem Komarichev}$^{1}$ \quad
  \textbf{Peter Wonka}$^{3}$ \quad
  \textbf{Evgeny Burnaev}$^{1,4}$ \\[6pt]
  \normalfont
  $^{1}$Applied AI Institute \\
  $^{2}$T-Tech \\
  $^{3}$KAUST \\
  $^{4}$AXXX
}

\begin{document}

\maketitle

\begin{abstract}
  Sparse-view novel view synthesis is a central problem in 3D content creation, but diffusion-based approaches remain limited by iterative denoising, making multi-view generation expensive at inference time. We introduce NAMVIS, a diffusion-free framework that reformulates multi-view image synthesis as geometry-conditioned next-scale autoregression. Instead of generating target views through repeated denoising, NAMVIS predicts discrete visual tokens through a small number of coarse-to-fine scale steps, while sampling all tokens within each scale and across target views in parallel. To anchor this generation process to explicit camera geometry, we propose Multi-scale Projective Pose Encoding, which injects source and target camera transformations into both target-view self-attention and source-to-target cross-attention at every resolution. NAMVIS further combines global conditioning with dense geometry-aware cross-attention, enabling the model to preserve source-view appearance while maintaining target-view consistency. Across Objaverse, GSO, and OmniObject3D, NAMVIS outperforms diffusion-based baselines in PSNR, SSIM, and LPIPS, while running over 3× faster than the evaluated diffusion baselines under the same evaluation setting. These results suggest that geometry-conditioned next-scale autoregression is a promising and efficient alternative to diffusion for sparse-view multi-view synthesis. Additional qualitative results, videos, and resources are available at \href{https://corl-team.github.io/namvis/}{corl-team.github.io/namvis}.
\end{abstract}

\begin{figure*}[!t]
    \centering
    \includegraphics[width=\textwidth]{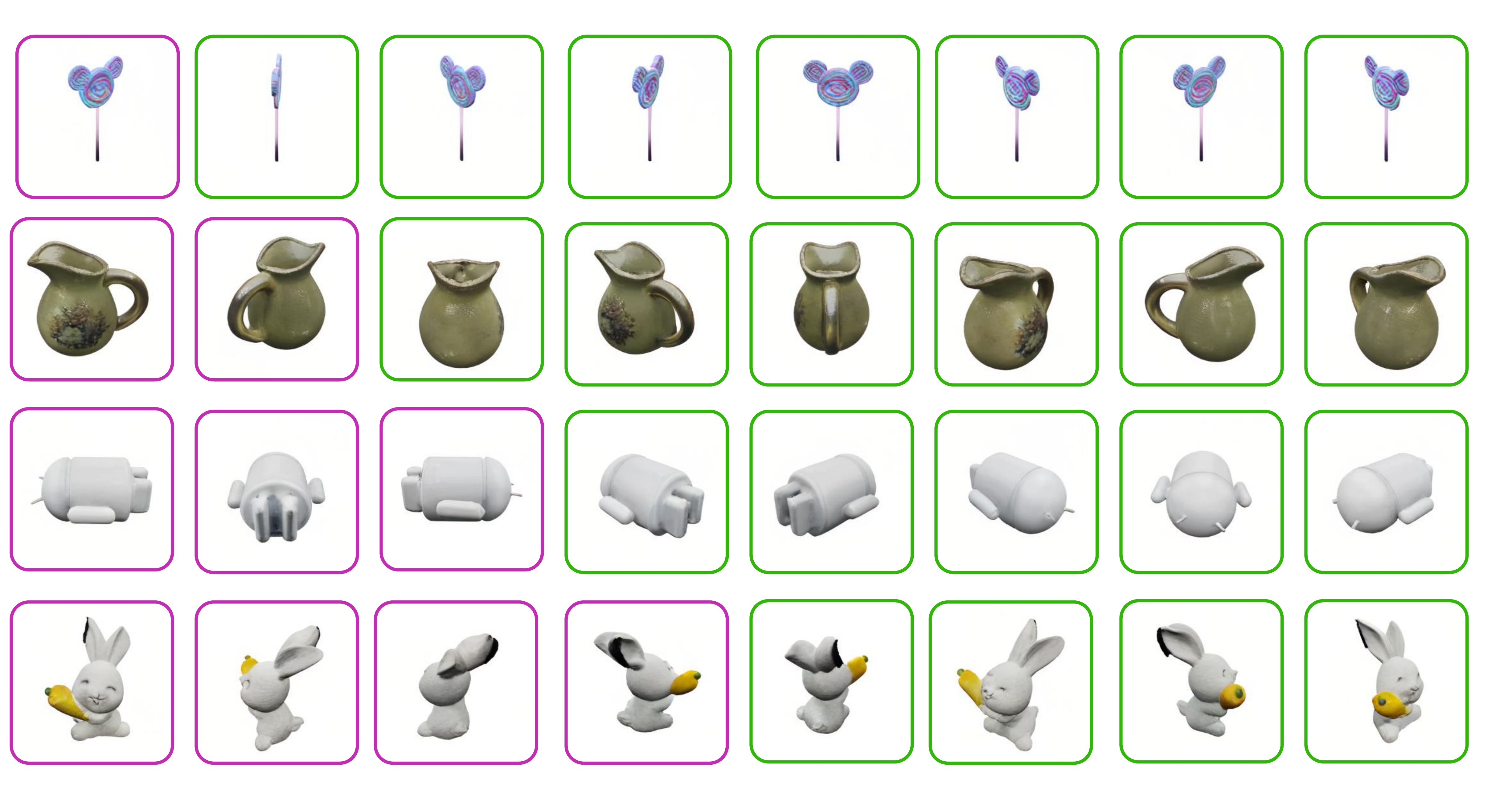}
    \caption{
    \textbf{NAMVIS} performs sparse-view novel view synthesis from posed source 
    images. Purple borders indicate input views; green borders indicate generated 
    target views. NAMVIS generates target views with next-scale 
    autoregressive decoding, producing sharp and more geometrically consistent 
    results.
    }
    \label{fig:teaser}
    \vspace{-0.5cm}
\end{figure*}

\section{Introduction}
\label{sec:intro}

Zero-shot novel view synthesis asks: given one or a few posed images of an unseen object, generate photorealistic images from new viewpoints—without per-scene optimization. The task is central to augmented reality, robotics, and 3D content creation~\cite{poole2022dreamfusion,tang2023make}, yet it is heavily under-constrained: a single photograph admits infinitely many consistent 3D scenes, so the model must supply strong priors over geometry, appearance, and illumination. In this work, we focus on posed sparse-view synthesis: given one or more source images with known cameras and a set of target cameras, the goal is to generate the corresponding target views without per-scene optimization or explicit 3D reconstruction.

Early methods reconstructed explicit 3D representations—NeRFs~\cite{mildenhall2021nerf}, 3D Gaussian Splatting~\cite{kerbl20233d}, or meshes~\cite{yariv2021volume,wang2021neus}—and required dozens to hundreds of views.
These representations collapse in the few-image regime due to depth and shape ambiguities~\cite{yu2021pixelnerf,chan2022efficient}. Zero-1-to-3~\cite{liu2023zero} showed that large pretrained 2D diffusion models~\cite{rombach2022high} already encode strong 3D priors: fine-tuning Stable Diffusion with relative pose conditioning enabled zero-shot NVS for the first time, spawning a wave of diffusion-based approaches~\cite{shi2023zero123++,qian2023magic123,long2024wonder3d,shi2023mvdream,liu2023syncdreamer,gu2023nerfdiff}.

While diffusion-based methods have achieved strong visual quality, their iterative denoising process can make multi-view inference expensive, especially when generating several target views.

Meanwhile, next-scale autoregressive models have matched or surpassed diffusion quality in text-to-image generation~\cite{tian2024visualautoregressivemodelingscalable,han2025infinityscalingbitwiseautoregressive}, yet their potential for 3D-aware synthesis remains largely untapped. 

In this paper, we introduce \textit{NAMVIS}, a purely autoregressive framework for zero-shot novel view synthesis from sparse observations. NAMVIS predicts target views through a denoising-free coarse-to-fine autoregression over discrete visual tokens, conditioned on a variable number of source images, their camera poses, and the desired target poses.

Our main contributions are:
\begin{itemize}
    \item We introduce NAMVIS, a next-scale autoregressive framework for sparse-view multi-view synthesis that predicts target views over coarse-to-fine scale blocks, avoiding iterative denoising.
    \item We propose Multi-scale Projective Pose Encoding, which extends projective camera-based pose encoding to the varying spatial resolutions of next-scale autoregression and injects geometry into both self-attention and source-to-target cross-attention.
    \item We design a dual-path conditioning mechanism that combines global source-view conditioning with dense geometry-aware cross-attention to preserve both semantic appearance and fine spatial detail.
    \item We evaluate NAMVIS on Objaverse, GSO, and OmniObject3D against representative diffusion-based baselines, showing improved PSNR, SSIM, LPIPS, and faster inference under the same evaluation setting.
    \item To support reproducibility and future research, we release the filtered scene list, rendered RGBA images, camera poses, masks, point maps, captions, and preprocessing scripts for approximately 200K objects.
\end{itemize}

\section{Related work}
\label{sec:related}

\paragraph{Optimization-based and feed-forward novel view synthesis.}

A useful axis for organizing NVS research is the \emph{generation paradigm}: methods that iterate (optimization-based or diffusion-based) versus methods that decode in a single pass (feed-forward or autoregressive).
Classical NVS reconstructs an intermediate 3D representation—a NeRF~\cite{mildenhall2021nerf,barron2022mip}, voxel grid~\cite{sitzmann2019deepvoxels}, or mesh~\cite{yariv2021volume,wang2021neus}—and renders novel views from it.
These methods excel when dozens of views are available but require per-scene optimization.

To bypass per-scene optimization, recent sparse-view reconstruction methods such as 
LRM~\cite{hong2024lrmlargereconstructionmodel} and LucidFusion~\cite{he2024lucidfusion} use feed-forward 
networks to predict explicit 3D representations, such as triplanes or 3D Gaussians, 
from sparse inputs. These methods address a closely related problem but differ in output space and inference objective: they first reconstruct an explicit 3D representation from which arbitrary views can be rendered, whereas NAMVIS directly synthesizes posed target images without producing a mesh, radiance field, or Gaussian representation. We therefore discuss feed-forward reconstruction methods as related work, while focusing our main experimental comparison on generative NVS methods that directly predict target views.

\paragraph{Diffusion-based generative NVS.}

A dominant generative NVS paradigm conditions pretrained 2D diffusion models on camera pose to implicitly capture 3D geometry.
Zero-1-to-3~\cite{liu2023zero} pioneered this direction by fine-tuning Stable Diffusion~\cite{rombach2022high} with relative pose conditioning.
Subsequent work improved multi-view consistency through cross-attention~\cite{qian2023magic123,shi2023zero123++}, synchronized multi-view generation~\cite{liu2023syncdreamer}, and joint prediction of multi-view appearance and geometry cues~\cite{long2024wonder3d}.
More recently, EscherNet~\cite{kong2024eschernetgenerativemodelscalable} introduced camera positional encodings to handle arbitrary numbers of reference and target views.
Despite strong generation quality, diffusion-based NVS methods typically rely on iterative denoising at inference, which increases latency when generating multiple target views.

\paragraph{Autoregressive visual generation.}

Autoregressive (AR) models factorize image distributions as sequential token predictions.
Early work~\cite{van2016conditional,van2017neural} operated in raster-scan order over discrete tokens.
VAR~\cite{tian2024visualautoregressivemodelingscalable} replaced token-by-token prediction with coarse-to-fine \emph{next-scale} prediction, improving speed and quality.
Infinity~\cite{han2025infinityscalingbitwiseautoregressive} scaled this paradigm further with continuous-discrete bitwise representations. 
Recently, autoregressive formulations have been explored for multi-view generation. Notably, CausNVS~\cite{kong2025causnvs} frames multi-view synthesis as an autoregressive sequence of \emph{diffusion} steps, but still relies on iterative denoising for the generation of each individual view.
NAMVIS bridges the gap between these domains by extending the \emph{discrete} next-scale AR paradigm to 3D-aware multi-view generation. By injecting explicit 3D camera geometry via multi-scale ProPE and geometry-modulated cross-attention, NAMVIS provides a diffusion-free autoregressive alternative to diffusion-based sparse-view multi-view synthesis.

\section{Method}

We address zero-shot novel view synthesis. Given source RGB images $I^{\text{src}}$ with camera poses $P^{\text{src}} \in SE(3)$ and target camera poses $P^{\text{tgt}} \in SE(3)$, we generate target images
\[
\hat{I}^{\text{tgt}} = F_\theta(I^{\text{src}}, P^{\text{src}}, P^{\text{tgt}}).
\]
NAMVIS instantiates $F_\theta$ as a next-scale autoregressive transformer built on Infinity~\cite{han2025infinityscalingbitwiseautoregressive}, which extends Visual Autoregressive (VAR) modeling~\cite{tian2024visualautoregressivemodelingscalable}.
We first review autoregressive next-scale models, then describe how we extend this design for pose-conditioned multi-view synthesis.

\subsection{Background: Autoregressive Next-Scale Models}

The core idea behind next-scale autoregression is to generate an image coarse-to-fine rather than pixel-by-pixel.
A multi-scale VQ-VAE encodes an image $I$ into a continuous feature map $f \in \mathbb{R}^{h \times w \times d}$ and $K$ residual token maps $(r_1, \dots, r_K)$, $r_k \in \mathbb{R}^{h_k \times w_k \times d}$, whose spatial resolution increases with $k$.
In VAR~\cite{tian2024visualautoregressivemodelingscalable}, each residual is vector-quantized to a discrete codebook index; the continuous feature map is recovered by accumulating upsampled residuals:
\[
f_k = \sum_{i=1}^k \operatorname{up}(r_i, (h, w)), \qquad f = f_K.
\]
A transformer predicts these indices scale by scale, conditioned on an external signal $c$ (\eg, a class label):
\[
p(r_{1:K} \mid c) = \prod_{k=1}^K p(r_k \mid r_{<k}, c).
\]

Infinity~\cite{han2025infinityscalingbitwiseautoregressive} replaces codebook-based quantization with a bitwise representation: each token is quantized to a binary code $r_k(m,n) \in \{-1,+1\}^{d}$, creating an implicit vocabulary of size $2^d$.
The transformer predicts each bit independently via $d$ binary classifiers, and a bitwise self-correction strategy stabilizes training.

\subsection{NAMVIS}

\begin{figure*}[t]
    \centering
    \vspace{-1cm}
    \includegraphics[trim=0 0 0 100, clip, width=\textwidth]{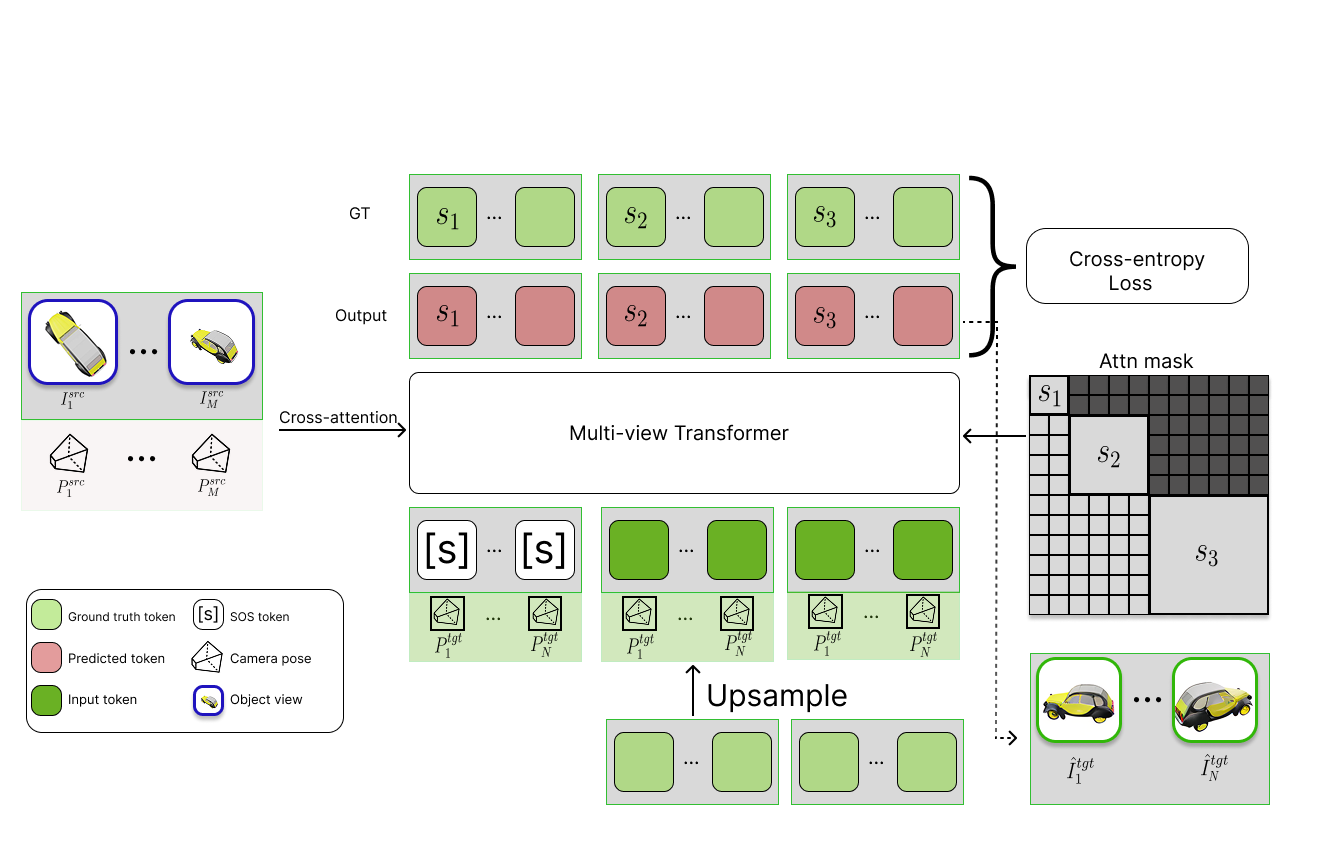}
    \caption{\textbf{NAMVIS architecture overview.}
    Source images are encoded by a frozen VQ-VAE into source features, which are pooled into a start-of-sequence token \texttt{[SOS]} and also serve as key--value pairs for dense cross-attention.
    The autoregressive transformer predicts target residual tokens scale by scale, with multi-scale ProPE injected into both self-attention (among target views) and cross-attention (from target to source). The frozen VQ-VAE decoder maps the final accumulated features to images.
    }
    \label{fig:architecture}
    \vspace{-0.5cm}
\end{figure*}

We show that multi-view image synthesis can be cast as geometry-conditioned next-scale prediction, eliminating the need for iterative denoising while preserving multi-view communication across target views. The key intuition is: if a coarse-to-fine autoregressive model can synthesize a photorealistic image from a text prompt, then we can replace the text conditioning with (i)~visual features from reference views and (ii)~3D camera geometry, steering the same generative process to produce novel views. Figure~\ref{fig:architecture} shows the overall pipeline; we describe each component below.

\paragraph{Training setup.}
During training, we sample random reference and target views from the same scene:
source images $I^{\text{src}}$ and target images $I^{\text{tgt}}$ with camera poses $P^{\text{src}}$ and $P^{\text{tgt}}$.
A shared, frozen VQ-VAE encoder maps both into continuous feature maps $f^{\text{src}}, f^{\text{tgt}} \in \mathbb{R}^{h \times w \times d}$.
We compute multi-scale residual token maps for the targets only:
$(r^{\text{tgt}}_1,\dots,r^{\text{tgt}}_K)$, $r^{\text{tgt}}_k \in \mathbb{R}^{h_k \times w_k \times d}$,
with the continuous feature at scale $k$ reconstructed as
$f^{\text{tgt}}_k = \sum_{i=1}^k \operatorname{up}(r^{\text{tgt}}_i, (h, w))$.

\paragraph{Autoregressive objective.}
The transformer predicts target residuals conditioned on source images and camera poses:
\[
p_\theta(r^{\text{tgt}}_{1:K} \mid f^{\text{src}}, P^{\text{src}}, P^{\text{tgt}})
= \prod_{k=1}^K p_\theta\!\big(r^{\text{tgt}}_k \mid r^{\text{tgt}}_{<k}, f^{\text{src}}, P^{\text{src}}, P^{\text{tgt}}\big).
\]
We retain Infinity's bitwise formulation and optimize a scale-reweighted cross-entropy loss over the independent bit predictions.
At inference, NAMVIS performs a denoising-free coarse-to-fine decode over $K$ scale steps: tokens within each scale and across target views are predicted in parallel, while later scales condition on earlier scales.
To enable classifier-free guidance (CFG), we drop the source conditioning with probability $10\%$ during training, replacing source features with a learned unconditional embedding.

\paragraph{Multi-view sequence layout and masking.}
To generate $N$ target views concurrently, we interleave token sequences scale by scale.
The combined sequence is $\mathcal{S} = (\mathcal{S}_1, \dots, \mathcal{S}_K)$, where $\mathcal{S}_k$ contains the flattened tokens of all $N$ views at scale $k$.
A block-causal attention mask enforces the autoregressive property: tokens in $\mathcal{S}_k$ attend to all tokens in $\mathcal{S}_{\leq k}$ (across all views) but are masked from $\mathcal{S}_{>k}$.
Within a scale block, tokens from different views attend to each other freely, enabling multi-view communication at every resolution.
A learned scale-level embedding $e_k \in \mathbb{R}^d$ is added before the first transformer block so that the model can distinguish the current resolution.

\paragraph{Multi-scale Projective Pose Encoding.}
Standard 2D RoPE captures image-plane position but does not encode the 3D relationship between cameras.
We therefore use a multi-scale projective positional encoding based on ProPE~\cite{li2025camerasrelativepositionalencoding}, adapted to next-scale autoregressive multi-view generation.

For a token feature $x_s$ at autoregressive scale $s$, we split each attention head into one projective part and two local rotary parts:
\[
x_s = [x^{\mathrm{proj}}_s,\; x^u_s,\; x^v_s].
\]
For view $v$ and scale $s$, our projective-RoPE transform is
\[
\Psi^{v,s}_b(x_s)
=
\left[
M_b(P_v)x^{\mathrm{proj}}_s,\;
\mathrm{RoPE}(x^u_s,u_s),\;
\mathrm{RoPE}(x^v_s,v_s)
\right],
\]
where $(u_s,v_s)$ are the token-grid coordinates at scale $s$, $P_v$ is the projective camera matrix for view $v$, and $b \in \{q,kv,o\}$ denotes the attention branch.
Following ProPE, the branch-specific projective matrices are
\[
M_q(P_v)=P_v^\top,\qquad
M_{kv}(P_v)=P_v^{-1},\qquad
M_o(P_v)=P_v.
\]
Thus, the projective component injects 3D camera geometry, while the two RoPE components preserve local 2D layout at the current autoregressive resolution.
Because NAMVIS generates tokens over multiple scales, we construct separate coordinate grids and rotary coefficients for each target scale $s$.

We then use the same projective-RoPE operator in two attention routes.
For target self-attention, queries, keys, values, and outputs all use the target-view camera:
\[
Y^{\mathrm{self}}_{t,s}
=
\Psi^{t,s}_o
\left(
\mathrm{Attn}
\left(
\Psi^{t,s}_q(Q_{t,s}),
\Psi^{t,\leq s}_{kv}(K_{t,\leq s}),
\Psi^{t,\leq s}_{kv}(V_{t,\leq s})
\right)
\right).
\]
The block-causal next-scale mask is applied inside this attention: tokens at scale $s$ may attend to previous scales and to all target views at the current scale, but not to future scales.

For source-to-target cross-attention, target tokens are queries and source features are keys and values.
Therefore the query and output branches use the target camera, while the key/value branches use the source cameras:
\[
Y^{\mathrm{cross}}_{t,s}
=
\Psi^{t,s}_o
\left(
\mathrm{Attn}
\left(
\Psi^{t,s}_q(Q_{t,s}),
\Psi^{src}_{kv}(K_{src}),
\Psi^{src}_{kv}(V_{src})
\right)
\right).
\]
Here, source features remain at the dense VQ-VAE feature resolution rather than being downsampled to the current target scale.
This lets every target scale attend to high-resolution source evidence while preserving the coarse-to-fine autoregressive structure.

This role-aware routing makes geometry available in both forms of communication: target self-attention encourages consistency among generated views, while source-to-target cross-attention anchors generation to posed source observations. In target self-attention, every token belongs to a generated target view.

\paragraph{Image conditioning.}
Reference images enter the transformer through two complementary paths---global and dense---serving low-frequency semantic alignment and high-frequency spatial detail respectively.

\emph{Global path.}
A learnable query $q \in \mathbb{R}^{d}$ cross-attends to the source features $f^{\text{src}}$, producing a pooled vector:
\[
\texttt{[SOS]} = \operatorname{CrossAttn}(q, f^{\text{src}}).
\]
This vector is prepended to the target sequence as a start-of-sequence token and is also fed through a mapping network to produce scale and shift parameters for Adaptive Layer Normalization (AdaLN), modulating every transformer block.
When generating multiple target views, we duplicate the \texttt{[SOS]} token for each output sequence.

\emph{Dense path.}
A single pooled token cannot capture high-frequency detail, so every transformer block also performs dense cross-attention against the unpooled VQ-VAE features $f^{\text{src}}$.
This cross-attention is modulated by ProPE (as described above), so the transformer aggregates source features according to the geometric relationship between source and target views.
Crucially, while the target queries are processed in a progressive, multi-scale fashion, the source keys and values are maintained at their full, single-scale resolution. This asymmetry ensures that even the coarsest stages of autoregressive generation are strictly anchored by high-fidelity spatial details from the source views.

\section{Experiments}

\subsection{Implementation Details}

\paragraph{Datasets.}

We train our model on a curated subset of Objaverse-XL~\cite{deitke2023objaversexluniverse10m3d}. 
Since raw Objaverse-XL 
contains many assets with rendering failures, missing textures, low foreground 
coverage, and degenerate geometry, we apply a three-stage filtering pipeline.
First, we remove assets with failed renders, low foreground occupancy, missing 
or predominantly white textures, and severe geometry artifacts.
Second, we score 
rendered views using an aesthetic-quality model and discard assets with 
consistently low visual quality.
Third, we generate image captions for the 
remaining assets and remove samples with degenerate captions, such as empty, 
generic, or visually inconsistent descriptions.
This combines geometry-, 
visibility-, appearance-, and caption-based filtering.

The resulting dataset contains approximately 200K objects. 
For each object, we 
render 100 posed RGBA views and store the corresponding camera intrinsics, camera 
extrinsics, masks, point maps, and generated captions.
To support reproducibility and future research, we release the filtered
scene list, rendered RGBA images, camera poses, masks, point maps, captions,
and preprocessing scripts.

\paragraph{Architecture.} 
NAMVIS is built upon the Infinity transformer architecture. 
The underlying  multi-scale VQ-VAE is pre-trained and kept frozen during transformer training. 

To reduce inference cost, we use standard KV caching together with a global 
ProPE caching mechanism. Since NAMVIS injects multi-scale projective camera 
encodings into multiple attention blocks, recomputing the geometry-dependent 
ProPE transformations at every layer would introduce redundant overhead. We 
therefore compute the ProPE projection tensors once for each source--target 
camera configuration and autoregressive scale, cache them globally, and reuse 
them across transformer layers. This reduces the cost of geometry-aware 
attention without changing the model outputs.

\paragraph{Training details.}
We train only the autoregressive transformer while keeping the VQ-VAE frozen. 
The model is optimized with AdamW using a base learning rate of 
$1 \times 10^{-4}$, a global batch size of 512, and 20 warm-up epochs, followed 
by cosine learning-rate decay. Training is run for 500 epochs in total. During 
training, we drop source-view conditioning with probability $0.1$ to enable 
classifier-free guidance at inference time.

\paragraph{Evaluation metrics.}
We evaluate novel-view fidelity using PSNR and SSIM, where higher is better, 
and perceptual similarity using LPIPS, where lower is better. 
All metrics are 
computed between generated target views and the corresponding ground-truth 
renders at $256 \times 256$ resolution. 
To assess multi-view consistency, we 
additionally report COLMAP reconstructability, measured by the average number of 
reconstructed sparse points from the generated multi-view outputs.

\subsection{Baselines}

To evaluate the zero-shot novel view synthesis capabilities of NAMVIS, we compare 
against representative public diffusion-based NVS baselines that directly generate 
target views from posed source images. Zero-1-to-3~\cite{liu2023zero} and its 
scaled-up version Zero-1-to-3 XL~\cite{deitke2023objaversexluniverse10m3d} serve 
as foundational single-view baselines that synthesize a novel view using relative 
camera conditioning. We also compare against recent multi-view diffusion methods: 
Wonder3D~\cite{long2024wonder3d}, which predicts multi-view appearance and 
geometry cues; SyncDreamer~\cite{liu2023syncdreamer}, which synchronizes 
multi-view generation to improve consistency; and EscherNet~\cite{kong2024eschernetgenerativemodelscalable}, 
which uses camera positional encoding (CaPE) to support flexible target-view 
generation. We use the official pretrained weights for all baselines.

For single-view baselines, we evaluate each available source view independently 
and report the best result according to PSNR. 
Some baselines impose restrictions on the camera elevation. Wonder3D~\cite{long2024wonder3d} requires source views with $0^\circ$ elevation, whereas SyncDreamer~\cite{liu2023syncdreamer} uses views at $30^\circ$ elevation. Therefore, we evaluate methods under their supported camera protocols: Wonder3D is evaluated at $0^\circ$ elevation, while the remaining methods are evaluated at $30^\circ$ elevation.

\begin{table*}[t]
\centering
\caption{Quantitative comparison across Objaverse, GSO, and OmniObject3D.}
\label{tab:quantitative_main}
\resizebox{\textwidth}{!}{%
\begin{tabular}{l ccc ccc ccc c}
\toprule
\multirow{2}{*}{Method} 
& \multicolumn{3}{c}{Objaverse (In-Domain)} 
& \multicolumn{3}{c}{GSO (Zero-Shot)} 
& \multicolumn{3}{c}{OmniObject3D (OO3D)} 
& \multirow{2}{*}{Time (s) $\downarrow$} \\
\cmidrule(lr){2-4} \cmidrule(lr){5-7} \cmidrule(lr){8-10}
& PSNR $\uparrow$ & SSIM $\uparrow$ & LPIPS $\downarrow$ 
& PSNR $\uparrow$ & SSIM $\uparrow$ & LPIPS $\downarrow$ 
& PSNR $\uparrow$ & SSIM $\uparrow$ & LPIPS $\downarrow$ 
& \\
\midrule
Zero-1-to-3~\cite{liu2023zero} 
& 19.097 & 0.816 & 0.152 
& 15.614 & 0.749 & 0.257 
& 16.094 & 0.768 & 0.219 
& 2.6 \\

Zero-1-to-3 XL~\cite{deitke2023objaversexluniverse10m3d}
& 19.341 & 0.827 & 0.136 
& 15.936 & 0.747 & 0.236 
& 16.320 & 0.767 & 0.209 
& 2.6 \\

Wonder3D~\cite{long2024wonder3d}
& 15.801 & 0.821 & 0.209 
& 14.158 & 0.777 & 0.285 
& 14.290 & 0.786 & 0.243 
& 2.0 \\

SyncDreamer~\cite{liu2023syncdreamer}
& 19.368 & 0.826 & 0.156 
& 16.961 & 0.781 & 0.246 
& 16.766 & 0.793 & 0.207 
& 16.6 \\

EscherNet~\cite{kong2024eschernetgenerativemodelscalable} 
& 19.078 & 0.816 & 0.147 
& 18.878 & 0.799 & 0.156 
& 17.767 & 0.780 & 0.167 
& 2.1 \\

\midrule
NAMVIS 1B (Ours) 
& \textbf{22.485} & \textbf{0.861} & \textbf{0.091} 
& \textbf{21.715} & \textbf{0.843} & \textbf{0.111} 
& \textbf{21.098} & \textbf{0.845} & \textbf{0.104} 
& \textbf{0.6} \\
\bottomrule
\end{tabular}%
}
\vspace{-0.5cm}
\end{table*}

\subsection{Quantitative Results}

We evaluate our method on Objaverse~\cite{deitke2022objaverseuniverseannotated3d}, 
GSO~\cite{downs2022googlescannedobjectshighquality}, and 
OmniObject3D (OO3D)~\cite{wu2023omniobject3dlargevocabulary3dobject} at a 
resolution of $256 \times 256$. Quantitative results are summarized in 
Table~\ref{tab:quantitative_main}. We report metrics averaged across nine 
source--target view configurations: 1-to-1, 1-to-2, 1-to-3, 2-to-1, 2-to-2, 
2-to-3, 3-to-1, 3-to-2, and 3-to-3. For each object, source and target views are 
sampled from a fixed set of available posed views. Full per-configuration results 
are provided in Appendix~\ref{app:runtime_scaling}.

The exact evaluation scenes for OO3D, GSO, and Objaverse are listed in Appendix~\ref{app:evaluation_scenes}.

\paragraph{Perceptual Quality.}

NAMVIS consistently outperforms diffusion-based baselines across all three datasets. On the in-domain Objaverse benchmark, NAMVIS achieves 22.485 PSNR, 0.861 SSIM, and 0.091 LPIPS, improving over Zero-1-to-3, Zero-1-to-3 XL, Wonder3D, SyncDreamer, and EscherNet. The gains are especially pronounced in LPIPS, indicating that NAMVIS produces perceptually sharper and more faithful novel views. On the zero-shot GSO benchmark, NAMVIS reaches 21.715 PSNR, 0.843 SSIM, and 0.111 LPIPS, suggesting strong generalization beyond the Objaverse training distribution. NAMVIS also performs best on OmniObject3D, achieving 21.098 PSNR, 0.845 SSIM, and 0.104 LPIPS, suggesting that the learned autoregressive multi-view prior transfers to more realistic object scans.

\paragraph{Inference Efficiency.}

In addition to improving image-quality metrics, NAMVIS substantially reduces 
inference latency. In the 1-to-1 setting, NAMVIS requires 0.6 seconds per target 
view, while the fastest evaluated diffusion baseline, Wonder3D, requires 
2.0 seconds. Thus, NAMVIS is approximately $3.3\times$ faster than the fastest 
diffusion baseline under the same $256 \times 256$ evaluation resolution. 

\paragraph{View-count scaling.}

We further evaluate NAMVIS with up to six source and six target views on Objaverse in Table~\ref{tab:view_scaling}. NAMVIS supports 6→6 joint generation while achieving 26.12 PSNR, 0.9015 SSIM, and 0.0473 LPIPS, with 1.56 s inference time and 5.0 GB peak VRAM. This indicates that the proposed multi-view autoregressive formulation scales to larger source and target sets without excessive inference cost.

\begin{table}[t]
    \centering
    \caption{Scaling to larger source and target view counts on Objaverse.}
    \label{tab:view_scaling}
    \begin{tabular}{lccccc}
    \toprule
    Setup & LPIPS $\downarrow$ & SSIM $\uparrow$ & PSNR $\uparrow$
    & Time (s) $\downarrow$ & VRAM (GB) $\downarrow$ \\
    \midrule
    2$\rightarrow$6 & 0.0654 & 0.8845 & 24.49 & 1.53 & 4.3 \\
    6$\rightarrow$2 & 0.0457 & 0.9017 & 26.07 & 0.84 & 4.1 \\
    6$\rightarrow$6 & 0.0473 & 0.9015 & 26.12 & 1.56 & 5.0 \\
    \bottomrule
    \end{tabular}
\end{table}

\paragraph{Cross-dataset generalization.}
All methods degrade when moving from Objaverse to GSO and OmniObject3D, but 
NAMVIS degrades more gracefully than the evaluated baselines. This suggests that 
geometry-conditioned next-scale autoregression learns a transferable object-level 
multi-view prior under our evaluation protocol.

\paragraph{Multi-View Consistency and Reconstructability.}
Standard 2D metrics such as PSNR, SSIM, and LPIPS evaluate per-view fidelity, but often overlook cross-view inconsistencies. To address this, we follow SyncDreamer~\cite{liu2023syncdreamer} and evaluate geometric consistency through COLMAP-based reconstructability. We run COLMAP on the source and generated target views for all methods using identical settings. This metric serves as a proxy for multi-view geometric consistency, as inconsistent views hinder stable feature tracking, camera registration, and triangulation.

\begin{wraptable}{r}{6cm}
    \centering
    \scriptsize
    \caption{
    COLMAP reconstructability comparison. Higher is better.
    }
    \label{tab:colmap_reconstructability}
    \begin{tabular}{l c}
        \toprule
        Method & COLMAP Points $\uparrow$ \\
        \midrule
        Wonder3D~\cite{long2024wonder3d} & 172 \\
        SyncDreamer~\cite{liu2023syncdreamer} & 151 \\
        EscherNet~\cite{kong2024eschernetgenerativemodelscalable} & 214 \\
        \midrule
        NAMVIS 1B (Ours) & \textbf{221} \\
        \bottomrule
    \end{tabular}
    \vspace{-0.3cm}
\end{wraptable}

As shown in Table~\ref{tab:colmap_reconstructability}, NAMVIS reconstructs an average of 221 sparse points (counting failed reconstructions as zero), outperforming EscherNet (214), Wonder3D (172), and SyncDreamer (151). These results indicate that our jointly generated views support more stable feature matching and triangulation than existing multi-view diffusion baselines. We exclude Zero-1-to-3~\cite{liu2023zero} and Zero-1-to-3 XL~\cite{deitke2023objaversexluniverse10m3d} from this comparison, as they do not natively generate jointly conditioned multi-view sets.

\paragraph{Registration coverage.}

We additionally measure camera-registration success across 30 scenes using identical COLMAP settings and report results in Table~\ref{tab:colmap_coverage}. NAMVIS registers 162 of 240 views (67.5\%), compared with 145 of 240 (60.4\%) for EscherNet, and approaches the 70.0\% coverage obtained using ground-truth images.

\begin{table}[H]
    \centering
    \caption{COLMAP camera-registration coverage across 30 scenes.}
    \label{tab:colmap_coverage}
        \begin{tabular}{lcc}
        \toprule
        Method & Registered views & Coverage $\uparrow$ \\
        \midrule
        GT        & 168 / 240 & 70.0\% \\
        EscherNet & 145 / 240 & 60.4\% \\
        NAMVIS    & 162 / 240 & \textbf{67.5\%} \\
        \bottomrule
        \end{tabular}
\end{table}

In addition, we provide qualitative dense camera-trajectory videos on the
project page, where NAMVIS produces stable geometry and appearance across
smooth viewpoint changes.

\subsection{Qualitative Results}

\begin{figure}[t]
    \centering
    \includegraphics[trim=0 50 0 20, clip, width=\textwidth]{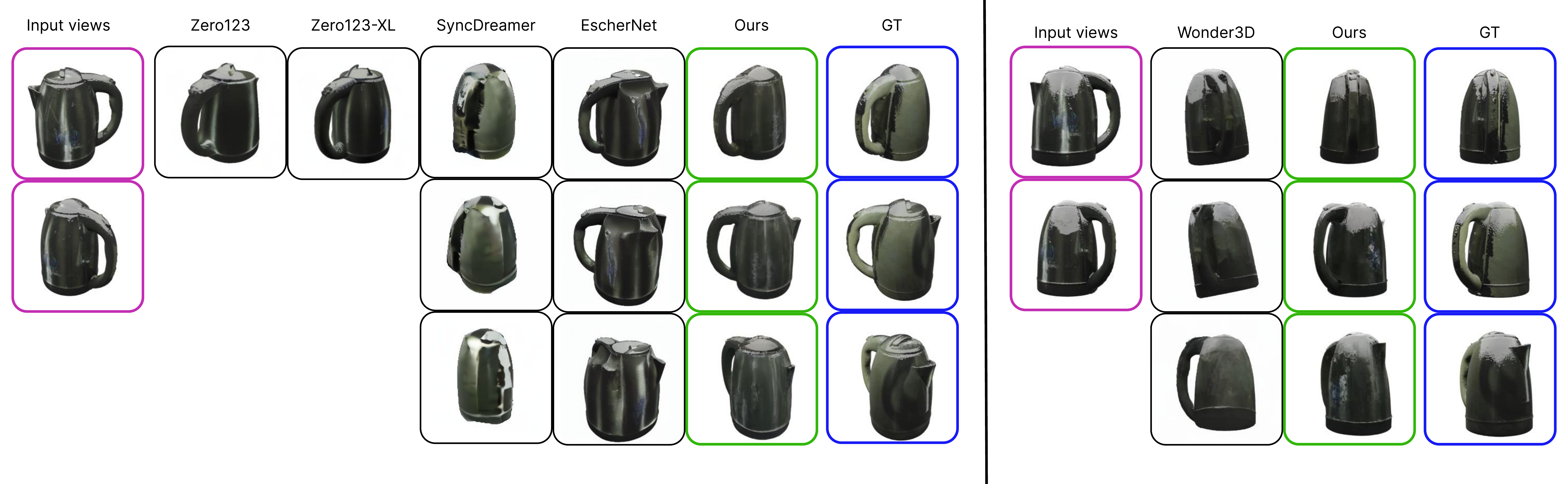}
    \caption{\textit{Qualitative comparison on  OO3D~\cite{wu2023omniobject3dlargevocabulary3dobject}.} }
    \label{fig:qualitative_comparison_oo3d1}
    \vspace{-0.3cm}
\end{figure}

\begin{figure}[t]
    \centering
    \includegraphics[trim=0 50 0 50, clip, width=\textwidth]{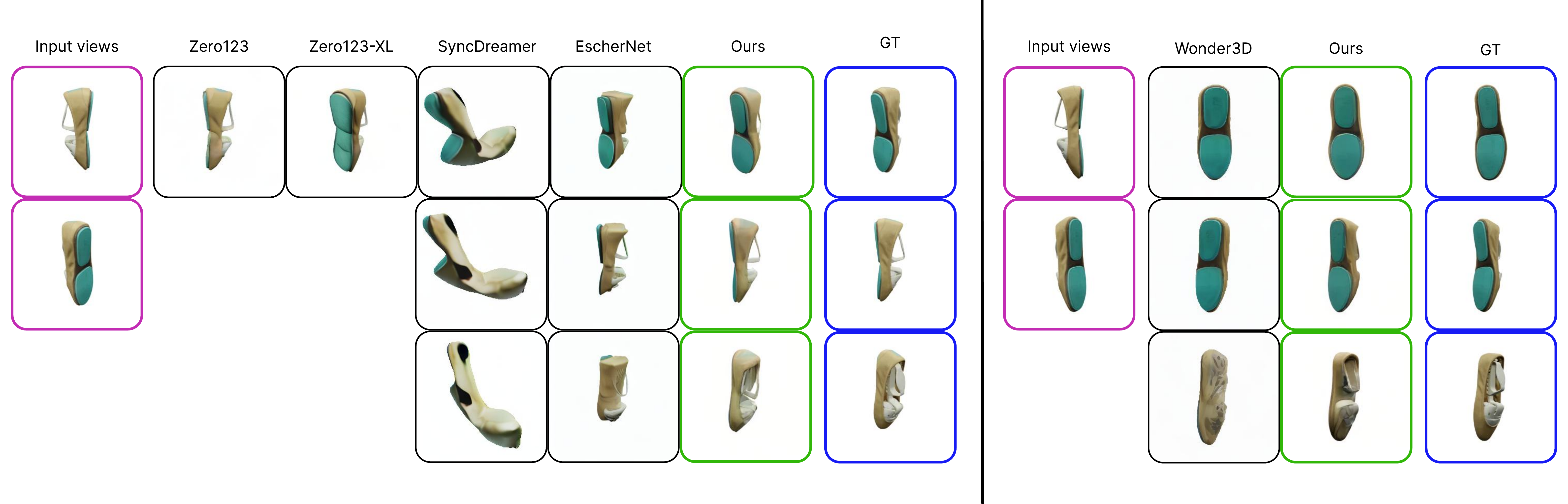}
    \caption{\textit{Qualitative comparison on  GSO~\cite{downs2022googlescannedobjectshighquality}.} }
    \label{fig:qualitative_comparison_gso1}
    \vspace{-0.5cm}
\end{figure}

Figures~\ref{fig:qualitative_comparison_oo3d1} and \ref{fig:qualitative_comparison_gso1} show qualitative comparisons on OO3D~\cite{wu2023omniobject3dlargevocabulary3dobject} and GSO~\cite{downs2022googlescannedobjectshighquality}. NAMVIS better preserves 
the source-view appearance while producing target views that follow the requested camera poses. To accommodate baseline constraints, Wonder3D~\cite{long2024wonder3d} is evaluated at $0^\circ$ camera elevation and all other baselines at $30^\circ$, while our method is evaluated at both angles. Compared with diffusion-based baselines, NAMVIS produces sharper object boundaries and more consistent geometry, especially for texture-rich and thin-structure objects.

Additional qualitative generations, source-view-count examples, and attention-map 
visualizations are provided in Appendices~\ref{app:additional_qualitative}, 
\ref{app:source_views}, and~\ref{app:attention_maps}.

\paragraph{Ablations.}

\begin{table}[!ht]
\footnotesize
\centering
\caption{
    \emph{Ablation study.} Each row replaces or removes one component of the full NAMVIS architecture. All variants are trained on a 10K subset of Objaverse-XL for 200 epochs.
}
\label{tab:ablation}
\begin{tabular}{l ccc}
\toprule
Model Variant & PSNR $\uparrow$ & SSIM $\uparrow$ & LPIPS $\downarrow$ \\
\midrule
\textit{Geometry \& Conditioning} & & & \\
w/ Pl\"ucker Rays (instead of ProPE)           & 20.260 & 0.899 & 0.142 \\
Prefix Conditioning (instead of Cross-Attn)    & 14.539 & 0.797 & 0.427 \\
\midrule
\textit{Feature Aggregation} & & & \\
CLIP \texttt{[SOS]} (instead of VQ-VAE)        & 20.613 & 0.901 & 0.126 \\
Learned Dense Features (instead of VQ-VAE)     & 20.886 & 0.901 & 0.134 \\
w/o Dense Cross-Attn (\texttt{[SOS]} only)     & 17.891 & 0.865 & 0.248 \\
\midrule
\textit{Optimization} & & & \\
LoRA Tuning (instead of Full Tuning)           & 18.912 & 0.883 & 0.185 \\
\midrule
\textbf{Full NAMVIS (Ours)}                    & \textbf{21.360} & \textbf{0.907} & \textbf{0.115} \\
\bottomrule
\end{tabular}%
\vspace{-0.5cm}
\end{table}

We conduct ablations on a 10K-object subset of our curated Objaverse-XL data. We ablate full-parameter fine-tuning as shown in Table~\ref{tab:ablation} by replacing it with LoRA~\cite{hu2021loralowrankadaptationlarge} with rank $r=256$. LoRA tuning performs substantially worse than full fine-tuning, suggesting that adapting a pretrained autoregressive image prior to geometry-conditioned multi-view synthesis requires updating the core transformer weights rather than only learning low-rank residual 
updates.

Additional results provided in Appendix~\ref{app:ablations} show that ProPE outperforms Pl\"ucker ray conditioning, dense cross-attention is critical for preserving fine spatial detail, and VQ-VAE source features outperform CLIP or separately learned dense 
features.

\subsection{Limitations and future work}

NAMVIS is trained and evaluated at \(256\times256\). Since the current checkpoint was trained only at this resolution, direct \(512\times512\) inference would require additional training or fine-tuning, as is common for transformer models trained at a fixed image resolution.

A promising direction is to combine NAMVIS with more efficient scale-wise transformer designs. Recent work such as Switti~\cite{voronov2025swittidesigningscalewisetransformers} suggests that next-scale generation may not require full causal attention to all previous scales, since the current-scale representation already contains accumulated lower-scale information. In NAMVIS, an analogous design would replace full block-causal attention over all previous target scales with scale-local or windowed target-target attention, while preserving geometry-aware source-to-target cross-attention. This could reduce memory and KV-cache cost at high resolutions without changing the core geometry-conditioned coarse-to-fine formulation.

Separately, our current model is trained on object-centric Objaverse-XL data, so its learned geometric priors may not transfer directly to complex, real-world scenes. However, the NAMVIS architecture itself is not restricted to object-centric inputs. Addressing this domain gap via larger, more diverse multi-view datasets that combine object-centric and scene-centric data is an important direction for future research.

\section{Conclusion}

We presented NAMVIS, a geometry-conditioned next-scale autoregressive framework for zero-shot novel view synthesis from sparse posed observations. Instead of generating target views through iterative diffusion denoising, NAMVIS performs coarse-to-fine autoregressive decoding over visual tokens, predicting all tokens within each scale and across target views in parallel. This provides an efficient 
diffusion-free alternative for sparse-view multi-view synthesis.

Across Objaverse, GSO, and OmniObject3D, NAMVIS improves PSNR, SSIM, and LPIPS over representative public diffusion-based baselines, including both single-view and multi-view generation methods. In the 1-to-1 setting, NAMVIS runs in 0.6\,s per target view, approximately $3.3\times$ faster than the fastest evaluated diffusion baseline under the same $256 \times 256$ resolution setting. 
Together with the ablation studies and COLMAP reconstructability evaluation, these results suggest that geometry-conditioned next-scale autoregression is a promising direction for efficient sparse-view novel view synthesis.

Future work will focus on reducing sequence-length bottlenecks for higher resolutions and denser target-view generation, improving robustness to pose noise, and extending training beyond object-centric data to more complex scene-level multi-view settings.

\bibliographystyle{plainnat}
\bibliography{neurips_2026}

\appendix

\section{Architectural Details}
\label{app:architectural_details}

NAMVIS uses a 1B-parameter transformer initialized from the Infinity~\cite{han2025infinityscalingbitwiseautoregressive} 2B checkpoint. 
To obtain the 1B model, we keep every other transformer block from the 2B model and discard the remaining blocks. Each retained transformer block contains self-attention followed by source-to-target cross-attention and an MLP. The final model has 16 transformer blocks, hidden dimension 2048, and 16 attention heads.

\section{Runtime and View-Count Scaling}
\label{app:runtime_scaling}

We evaluate NAMVIS for different numbers of source and target views. Table~\ref{tab:results_views} reports inference time and image-quality metrics for each source--target configuration, where setup $a$--$b$ denotes $a$ source views and $b$ target views. Inference time is measured at $256 \times 256$ resolution on a single NVIDIA A40 GPU.

Increasing the number of source views generally improves reconstruction quality, as reflected by lower LPIPS and higher SSIM and PSNR scores. This is expected, since additional source views reduce ambiguity in unobserved regions. Increasing the number of target views primarily affects computational cost: runtime increases as more target views are generated jointly, while image quality remains relatively 
stable for a fixed number of source views.

Overall, these results show that NAMVIS scales efficiently with the number of target views while benefiting from additional source-view evidence.

\begin{table*}[t]
\centering
\scriptsize
\setlength{\tabcolsep}{3.5pt}
\caption{Inference time and image quality for different source--target view configurations.
Setup $a$--$b$ denotes $a$ source views and $b$ target views.}
\label{tab:results_views}
\resizebox{\textwidth}{!}{
\begin{tabular}{llrrrrrrrr}
\toprule
\textbf{Dataset} & \textbf{Setup} &
\multicolumn{4}{c}{\textbf{NAMVIS}} &
\multicolumn{4}{c}{\textbf{EscherNet~\cite{kong2024eschernetgenerativemodelscalable}}} \\
\cmidrule(lr){3-6} \cmidrule(lr){7-10}
& &
\textbf{Time (s)} & \textbf{LPIPS$\downarrow$} & \textbf{SSIM$\uparrow$} & \textbf{PSNR$\uparrow$} &
\textbf{Time (s)} & \textbf{LPIPS$\downarrow$} & \textbf{SSIM$\uparrow$} & \textbf{PSNR$\uparrow$} \\
\midrule

\textbf{GSO}
& 1--1 & 0.60 & 0.1634 & 0.8104 & 19.0542 & 2.09 & 0.2924 & 0.7078 & 14.0517 \\
& 1--2 & 0.81 & 0.1668 & 0.8100 & 19.2163 & 2.16 & 0.2705 & 0.7334 & 14.5864 \\
& 1--3 & 0.94 & 0.1698 & 0.8110 & 19.1860 & 2.35 & 0.2683 & 0.7342 & 14.8349 \\
& 1--4 & 1.11 & 0.1633 & 0.8147 & 19.5188 & 2.78 & 0.2498 & 0.7498 & 15.6767 \\
\midrule
& 2--1 & 0.62 & 0.0810 & 0.8603 & 23.0655 & 2.08 & 0.0927 & 0.8360 & 21.2183 \\
& 2--2 & 0.79 & 0.0883 & 0.8538 & 22.4639 & 2.17 & 0.1037 & 0.8336 & 20.6660 \\
& 2--3 & 0.95 & 0.1048 & 0.8451 & 21.8491 & 2.44 & 0.1208 & 0.8200 & 19.8665 \\
& 2--4 & 1.11 & 0.1061 & 0.8452 & 21.9103 & 2.81 & 0.1182 & 0.8233 & 20.1216 \\
\midrule
& 3--1 & 0.63 & 0.0737 & 0.8675 & 23.5606 & 2.20 & 0.0890 & 0.8366 & 21.2918 \\
& 3--2 & 0.80 & 0.0725 & 0.8692 & 23.6664 & 2.23 & 0.0808 & 0.8488 & 21.8496 \\
& 3--3 & 0.96 & 0.0810 & 0.8632 & 23.3698 & 2.47 & 0.0890 & 0.8415 & 21.5421 \\
& 3--4 & 1.12 & 0.0835 & 0.8608 & 23.3741 & 2.84 & 0.0936 & 0.8393 & 21.5674 \\
\midrule
& 4--1 & 0.63 & 0.0721 & 0.8704 & 23.8080 & 2.25 & 0.0846 & 0.8382 & 21.5261 \\
& 4--2 & 0.82 & 0.0691 & 0.8712 & 23.9583 & 2.27 & 0.0778 & 0.8501 & 22.1110 \\
& 4--3 & 0.96 & 0.0678 & 0.8708 & 24.1201 & 2.52 & 0.0737 & 0.8511 & 22.3149 \\
& 4--4 & 1.12 & 0.0666 & 0.8726 & 24.3523 & 2.98 & 0.0722 & 0.8532 & 22.5996 \\

\midrule
\textbf{OO3D}
& 1--1 & 0.60 & 0.1505 & 0.8095 & 18.6618 & 2.09 & 0.2842 & 0.7151 & 13.6204 \\
& 1--2 & 0.81 & 0.1529 & 0.8098 & 18.4677 & 2.16 & 0.2772 & 0.7345 & 14.1137 \\
& 1--3 & 0.94 & 0.1644 & 0.8021 & 18.0196 & 2.35 & 0.2815 & 0.7367 & 13.9419 \\
& 1--4 & 1.11 & 0.1545 & 0.8091 & 18.4509 & 2.78 & 0.2627 & 0.7455 & 14.4499 \\
\midrule
& 2--1 & 0.62 & 0.0791 & 0.8609 & 22.2238 & 2.08 & 0.1254 & 0.8208 & 19.1116 \\
& 2--2 & 0.79 & 0.0851 & 0.8598 & 22.0147 & 2.17 & 0.1146 & 0.8353 & 19.3928 \\
& 2--3 & 0.95 & 0.0949 & 0.8513 & 21.3968 & 2.44 & 0.1244 & 0.8267 & 18.9607 \\
& 2--4 & 1.11 & 0.0907 & 0.8525 & 21.5572 & 2.81 & 0.1223 & 0.8301 & 19.1145 \\
\midrule
& 3--1 & 0.63 & 0.0721 & 0.8657 & 22.7968 & 2.20 & 0.1131 & 0.8287 & 19.5381 \\
& 3--2 & 0.80 & 0.0691 & 0.8742 & 23.2689 & 2.23 & 0.0937 & 0.8523 & 20.5691 \\
& 3--3 & 0.96 & 0.0717 & 0.8736 & 23.0275 & 2.47 & 0.0911 & 0.8548 & 20.6615 \\
& 3--4 & 1.12 & 0.0726 & 0.8701 & 22.9430 & 2.84 & 0.0946 & 0.8512 & 20.4794 \\
\midrule
& 4--1 & 0.63 & 0.0695 & 0.8693 & 23.1950 & 2.25 & 0.1079 & 0.8325 & 19.7964 \\
& 4--2 & 0.82 & 0.0659 & 0.8769 & 23.6277 & 2.27 & 0.0865 & 0.8560 & 21.0037 \\
& 4--3 & 0.96 & 0.0668 & 0.8781 & 23.5622 & 2.52 & 0.0823 & 0.8615 & 21.2054 \\
& 4--4 & 1.12 & 0.0657 & 0.8777 & 23.7146 & 2.98 & 0.0814 & 0.8610 & 21.3135 \\

\bottomrule
\end{tabular}
}
\end{table*}

\paragraph{Caching breakdown.}

Table~\ref{tab:cache_time} reports the effect of caching on inference time in the 3-to-3 setting. Standard KV caching reduces repeated attention computation across autoregressive scales, while ProPE caching avoids recomputing geometry-dependent projective transformations across transformer layers. Removing KV caching increases runtime from 0.96\,s to 1.06\,s, while removing ProPE caching 
increases runtime to 1.28\,s, indicating that caching geometry-dependent ProPE features is particularly important for efficient multi-view inference.

\begin{table}[t]
\centering
\small
\caption{Inference time for a 3-to-3 setup for different caching configurations.}
\label{tab:cache_time}
\setlength{\tabcolsep}{12pt}
\begin{tabular}{lc}
\toprule
\textbf{Setup} &
\textbf{Time (s)} \\
\midrule
NAMVIS & 0.96 \\
w/o KV-cache & 1.06 \\
w/o ProPE-cache & 1.28 \\
\bottomrule
\end{tabular}
\end{table}

\section{Effect of the Number of Source Views}
\label{app:source_views}

\begin{figure}[t]
    \centering
    \includegraphics[width=0.8\linewidth]{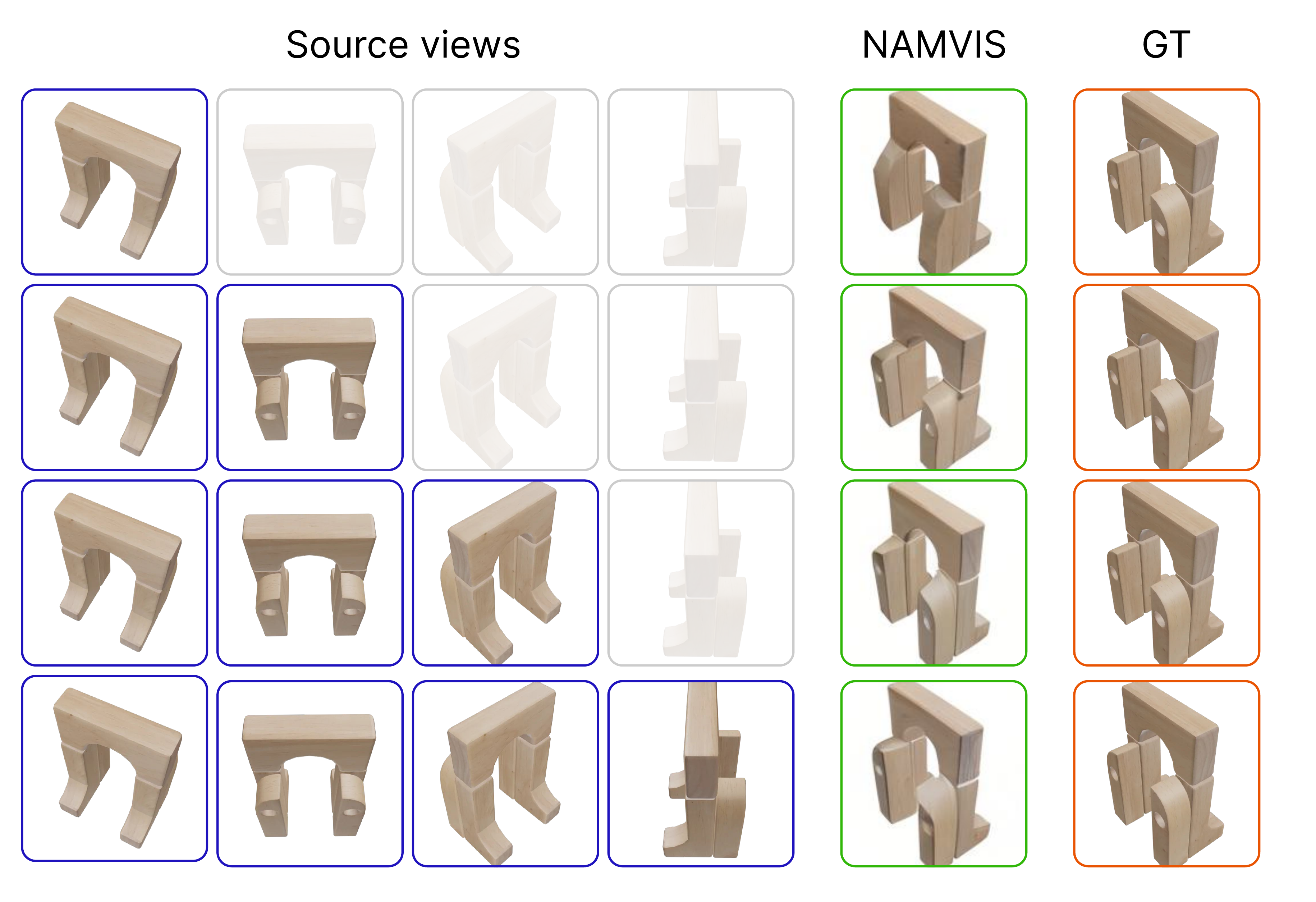}
    \caption{\textit{Varying number of source views from 1 to 4. When views covering previously unseen areas appear, the generated image quality improves.} }
    \label{fig:varying_src1}
\end{figure}

\begin{figure}[t]
    \centering
    \includegraphics[width=0.8\linewidth]{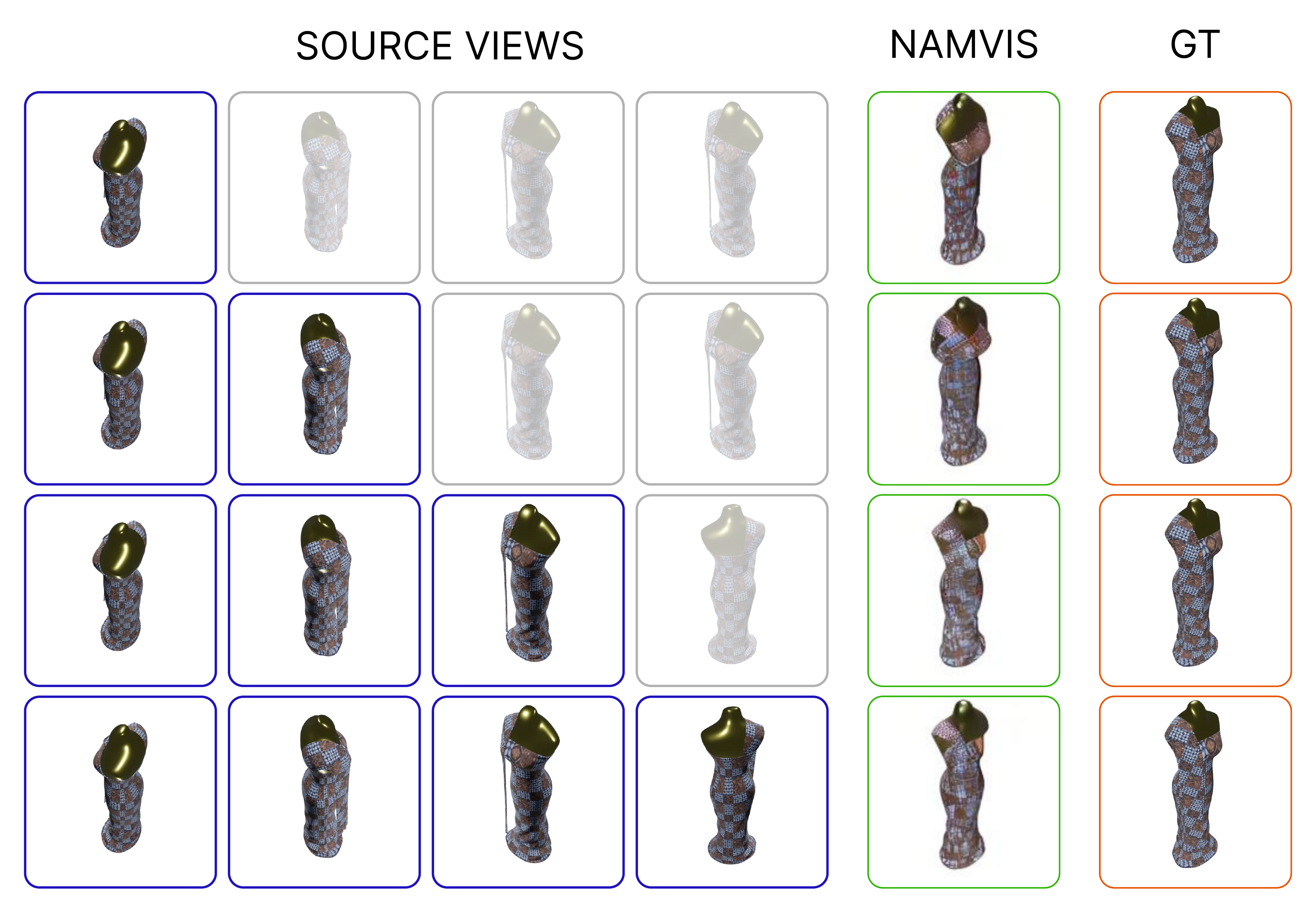}
    \caption{\textit{Varying number of source views from 1 to 4. When views covering previously unseen areas appear, the generated image quality improves.} }
    \label{fig:varying_src2}
\end{figure}

Figures~\ref{fig:varying_src1} and~\ref{fig:varying_src2} show how increasing 
the number of source views affects generation quality. As additional views become 
available, NAMVIS shifts from hallucinating unobserved regions to rendering 
object parts that are directly supported by the input views.

\section{Attention Maps Analysis}
\label{app:attention_maps}

\begin{figure*}[t]
    \centering
    \includegraphics[width=0.9\linewidth]{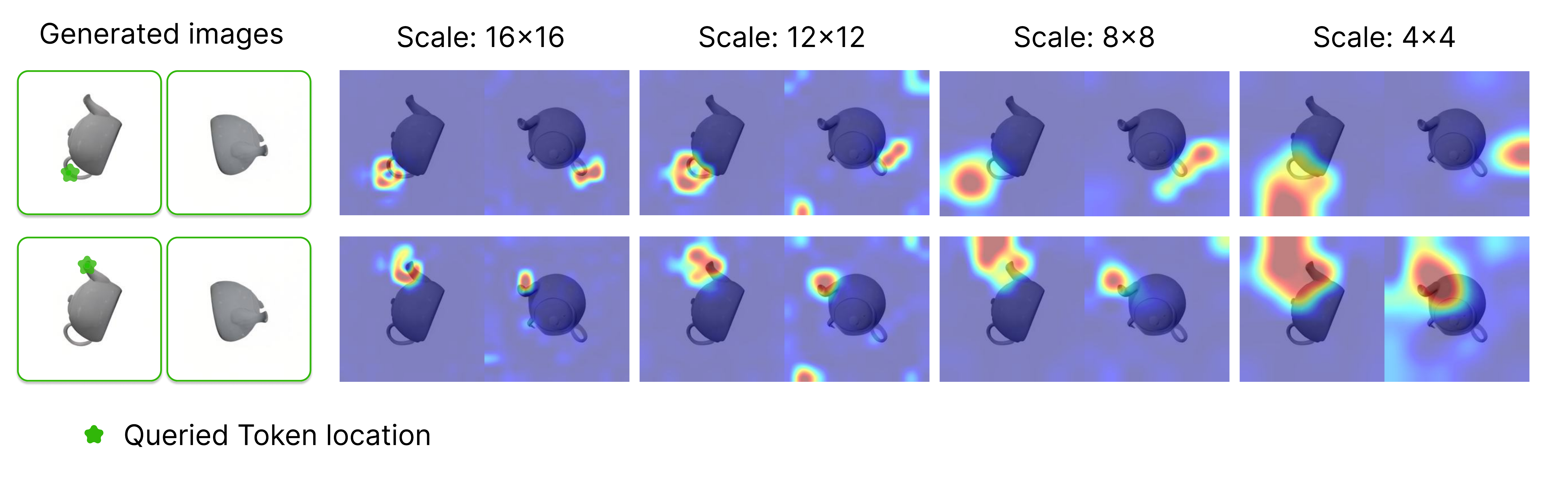}
    \caption{\textit{Target-view self-attention pattern between two generated views.} For a queried token in the first target view, self-attention places high weight on corresponding regions in both generated target views, suggesting that target-target attention supports multi-view consistency.}
    \label{fig:self_attention}
\end{figure*}

\begin{figure*}[t]
    \centering
    \includegraphics[width=0.9\linewidth]{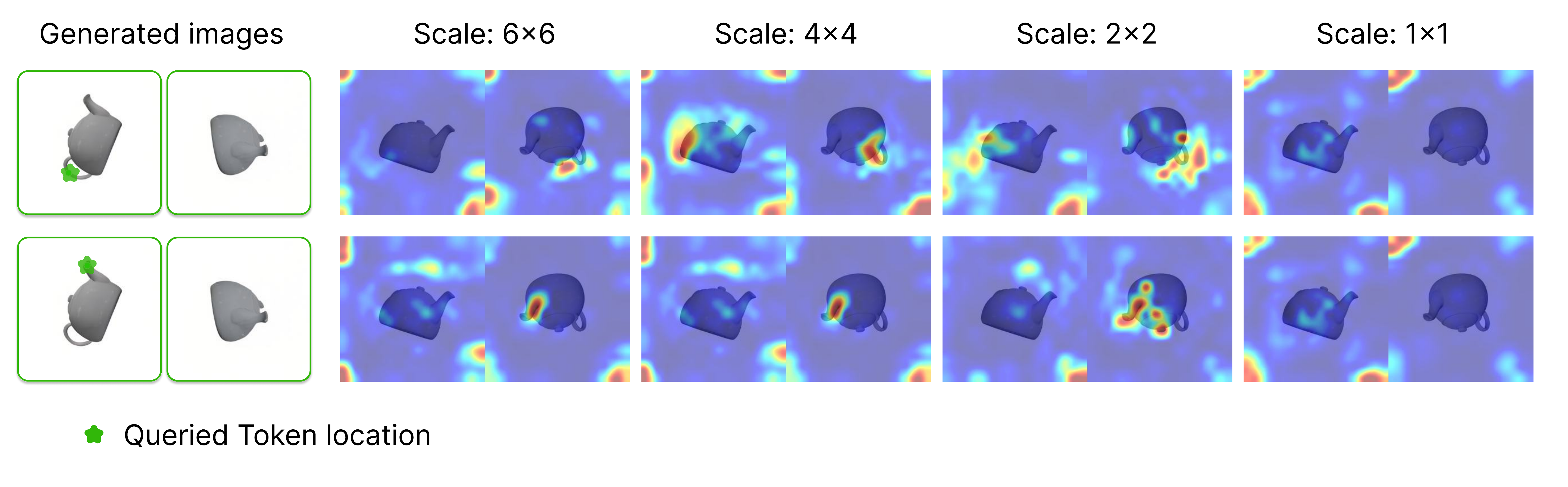}
    \caption{\textit{Cross-attention pattern from a target image to two source images, layer 14.} Cross-attention assigns high weights to source-image regions that are visually aligned with the queried target location.}
    \label{fig:cross_attention}
\end{figure*}

Visualizations of self-attention and cross-attention maps suggest that NAMVIS 
learns view-correspondence patterns consistent with the underlying camera 
geometry. For a query token in a generated target image, the model often assigns 
high attention weights to visually and geometrically corresponding regions across 
other views. We also observe a scale-dependent pattern: self-attention tends to 
emphasize target-target consistency at finer scales 
(Figure~\ref{fig:self_attention}), while cross-attention focuses on source-target 
alignment at coarser scales (Figure~\ref{fig:cross_attention}).

\section{Ablation Details}
\label{app:ablations}

\paragraph{Ablation studies.}
We conduct ablation studies to isolate the main architectural and optimization choices of NAMVIS, as reported in Table~\ref{tab:ablation}. Due to the cost of full-scale training, all ablation variants are trained on a randomly sampled 10K subset of our filtered Objaverse dataset for 200 epochs and evaluated on a held-out subset.  These experiments are intended to compare design choices under a fixed compute budget. 

\subsection{Camera Pose Parameterization}

Effective camera pose parameterization is critical for multi-view generation, as the generated images must strictly adhere to the target geometry. We compare our relative geometric approach against absolute pose conditioning via Pl\"ucker raymaps~\cite{zhang2024camerasraysposeestimation}. In the raymap baseline, explicitly computed 3D rays are concatenated directly along the feature dimension. As shown in Table \ref{tab:ablation},  injecting 3D camera geometry directly into the attention mechanism via multiscale ProPE~\cite{li2025camerasrelativepositionalencoding} outperforms the Pl\"ucker raymap conditioning across all metrics. This suggests that injecting geometry into the attention mechanism is more effective in our setting than concatenating explicit ray features.

\subsection{Reference Image Conditioning}

Our geometry-aware reference conditioning operates via two distinct mechanisms: globally modulating the transformer blocks and initializing the sequence using a pooled \texttt{[SOS]} token, and aggregating fine-grained spatial details via dense cross-attention. We 
ablate both the source of these conditioning signals and the way they are injected into the autoregressive transformer.

\paragraph{Global Initialization (\texttt{[SOS]} Token).}
We ablate the source of the global semantic conditioning by replacing our VQ-VAE attentive pooling with pre-trained CLIP features~\cite{radford2021learningtransferablevisualmodels}. While CLIP provides strong semantic priors, the VQ-VAE pooled features yield superior perceptual quality. Because the target sequence is composed of VQ-VAE tokens, deriving the \texttt{[SOS]} token from the same latent space provides better stylistic alignment and eliminates the domain gap between condition and target.

\paragraph{Dense Feature Aggregation.}
To understand the importance of high-frequency spatial conditioning, we completely remove the dense cross-attention, relying solely on the \texttt{[SOS]} token. This causes a substantial degradation in quality, suggesting that a single global token lacks the capacity to guide fine-grained image-to-image translation.

Next, we evaluate the injection mechanism itself. Replacing our dual-attention design (self-attention followed by cross-attention) with standard prefix conditioning, where source tokens are simply prepended to the target sequence, leads to a large performance drop. We hypothesize that directly prefixing dense source tokens interferes with the scale-structured autoregressive sequence, whereas cross-attention keeps source features in a separate key-value space and allows the target sequence to preserve its coarse-to-fine structure.

Finally, we ablate the source of the dense features by training a separate, dedicated feature extractor ("Learned Dense Features") via ConvNeXt~\cite{liu2022convnet2020s} instead of utilizing the frozen VQ-VAE representations. The frozen VQ-VAE not only avoids the computational overhead of an auxiliary encoder but also performs marginally better, further validating the efficiency of our shared latent space design.

\section{Additional Qualitative Results}
\label{app:additional_qualitative}

\begin{figure}[t]
    \centering
    \includegraphics[width=0.8\linewidth]{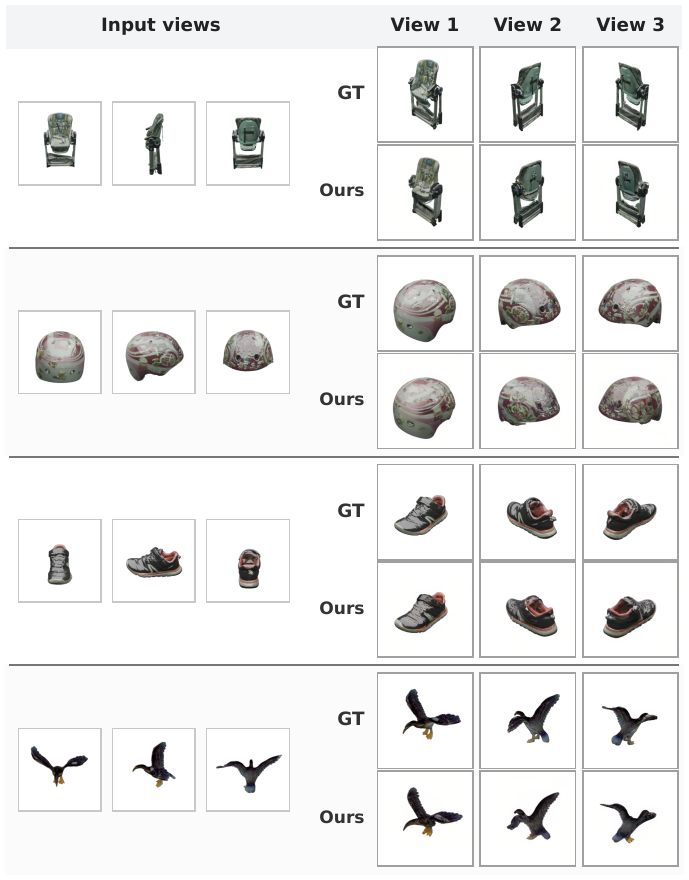}
    \caption{\textit{Qualitative results of our method on  OO3D~\cite{wu2023omniobject3dlargevocabulary3dobject}.} }
    \label{fig:qualitative_namvis1}
    \vspace{-0.5cm}
\end{figure}

\begin{figure}[t]
    \centering
    \includegraphics[width=0.8\linewidth]{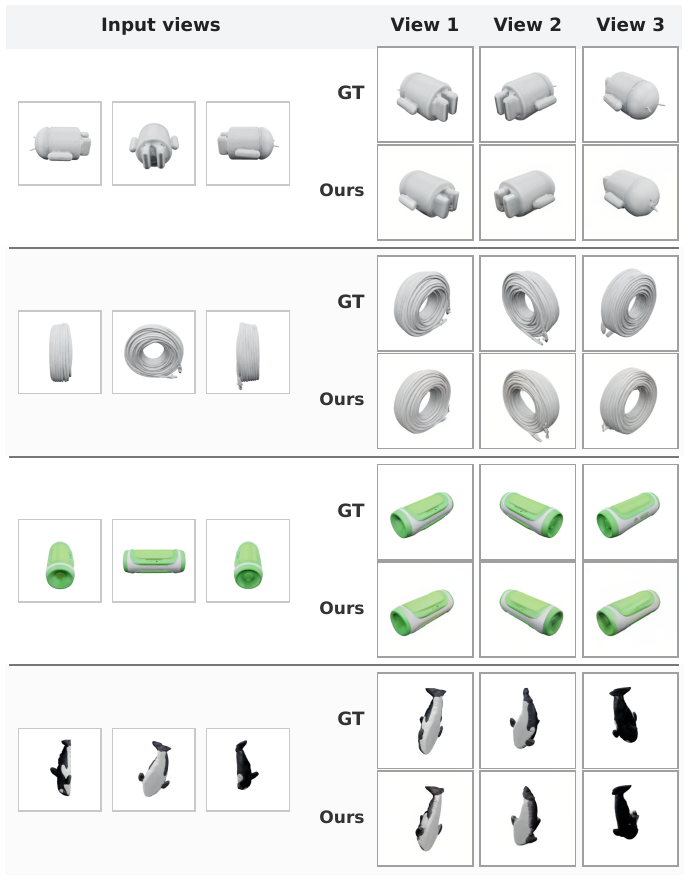}
    \caption{\textit{Qualitative results of our method on GSO~\cite{downs2022googlescannedobjectshighquality}.} }
    \label{fig:qualitative_namvis2}
    \vspace{-0.5cm}
\end{figure}

Figures~\ref{fig:qualitative_namvis1} and~\ref{fig:qualitative_namvis2} show 
additional NAMVIS generations across diverse object categories, shapes, poses, 
and textures on OO3D and GSO.

\paragraph{Scale progression.}

\begin{figure*}[ht]
    \centering
    \includegraphics[width=1.0\linewidth]{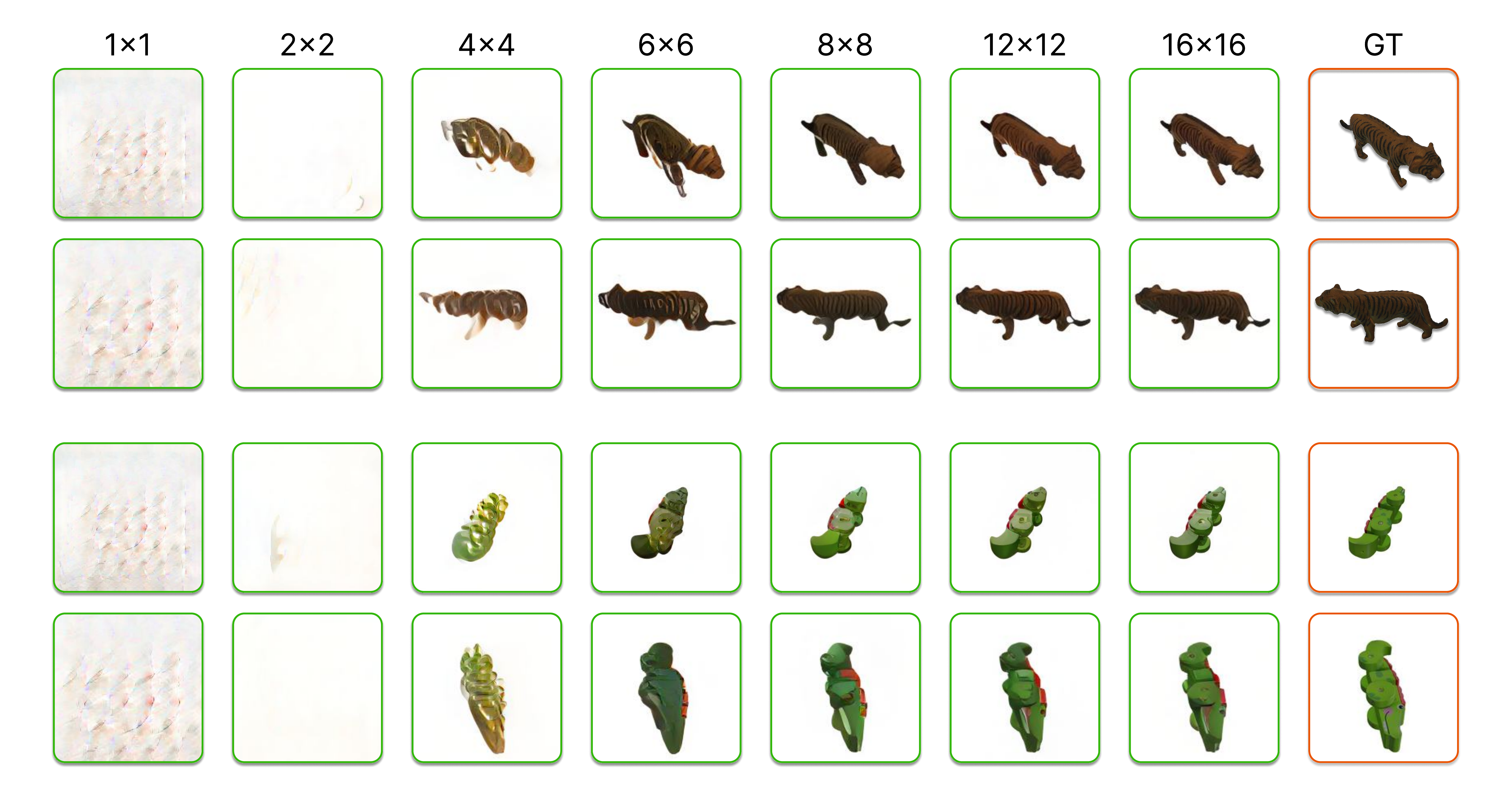}
    \caption{Visualization of image progression as new scales are summed up.}
    \label{fig:scale_progression}
\end{figure*}

In Figure \ref{fig:scale_progression}, we show how details appear on the final generated image as additional scales are summed up. Early scales $1\times 1$ and $2 \times 2$ do not contribute significantly, the real final image contours start appearing after scale $4 \times 4$ is added. The image then progressively accumulates finer details at later scales.

\section{Attention Maps Analysis}

\begin{figure*}[ht]
    \centering
    \includegraphics[width=1.0\linewidth]{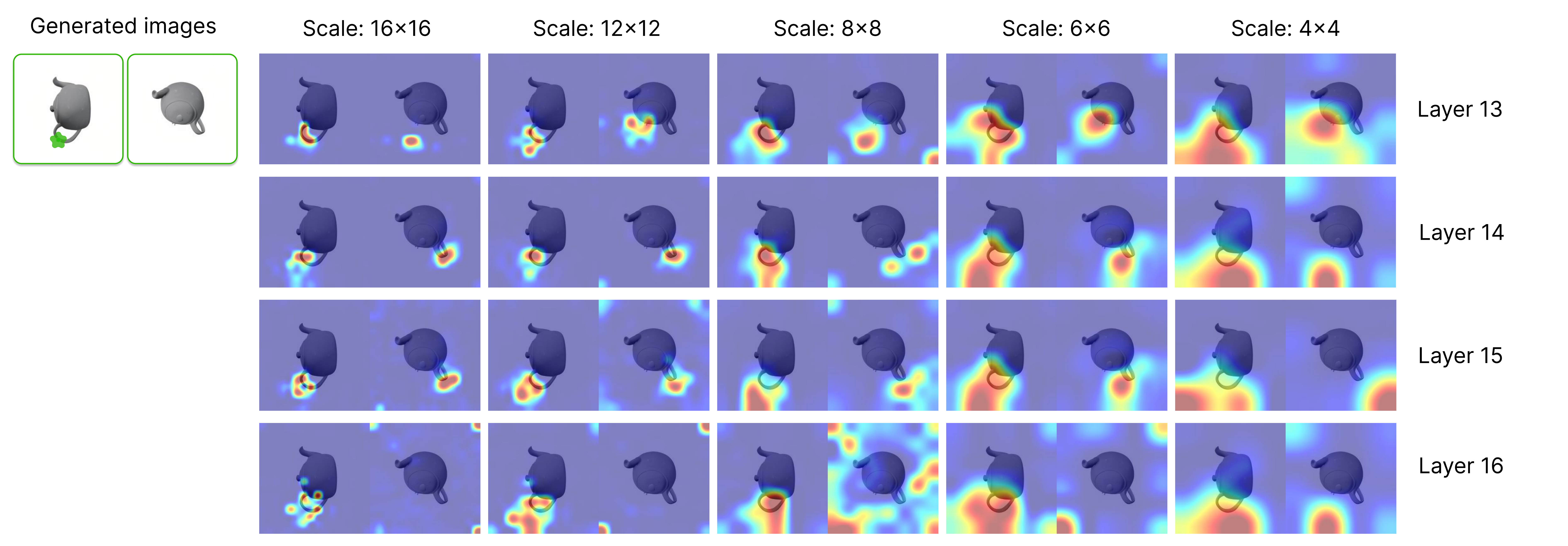}
    \caption{Visualization of self-attention pattern on layers 13-16, scales $4 \times 4$, $6 \times 6$, $8 \times 8$, $12 \times 12$, $16 \times 16$.}
    \label{fig:self_attn}
\end{figure*}

\begin{figure*}[ht]
    \centering
    \includegraphics[width=1.0\linewidth]{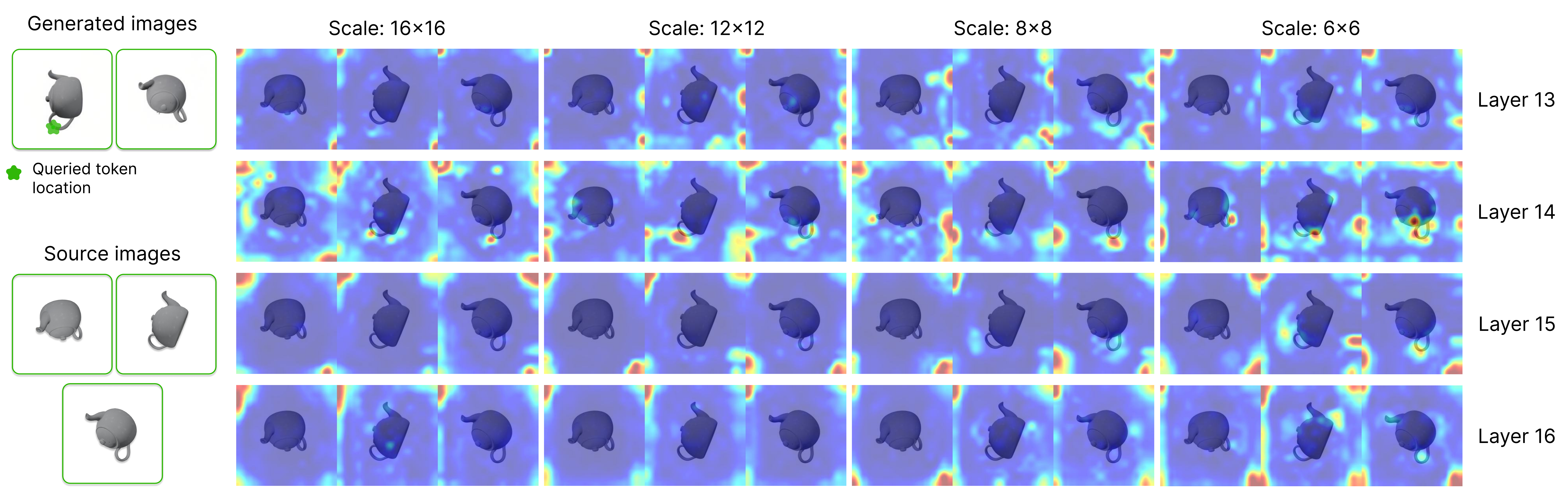}
    \caption{Visualization of cross-attention pattern on layers 13-16, scales $6 \times 6$, $8 \times 8$, $12 \times 12$, $16 \times 16$.}
    \label{fig:cross_attn1}
\end{figure*}

\begin{figure*}[ht]
    \centering
    \includegraphics[width=1.0\linewidth]{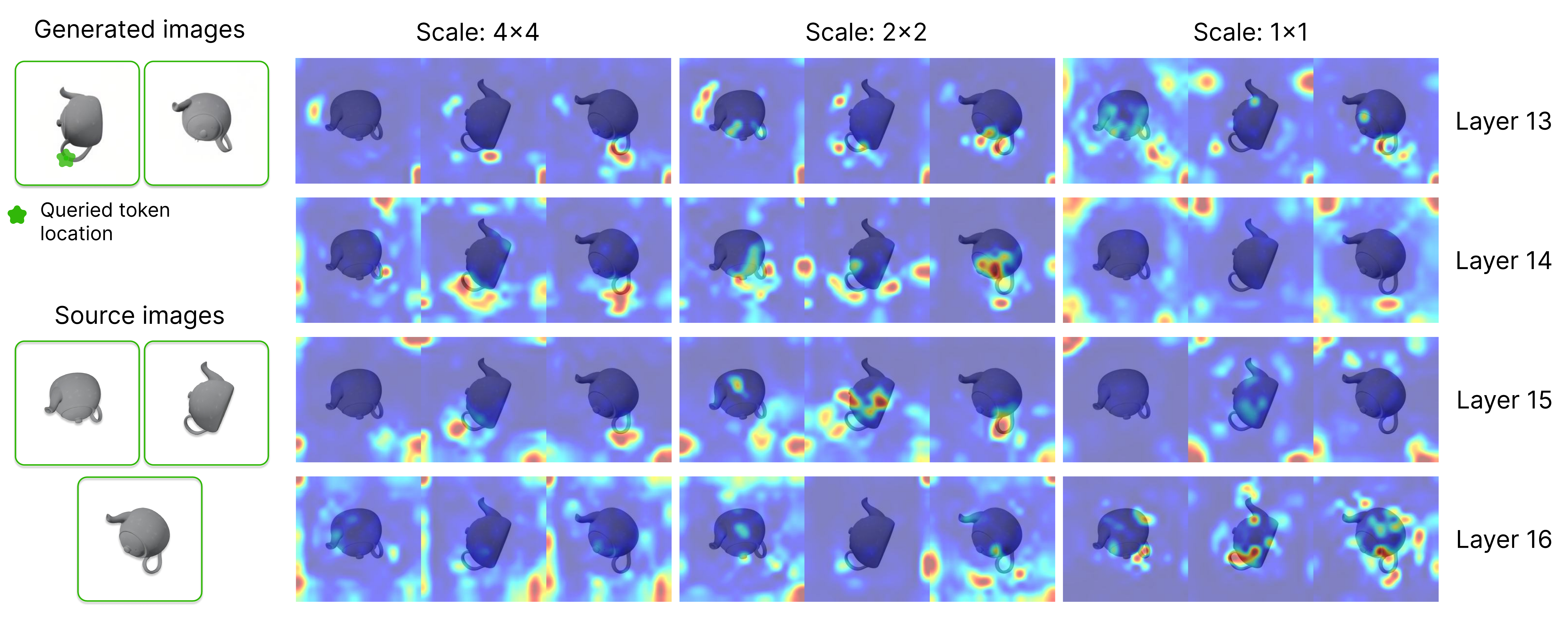}
    \caption{Visualization of cross-attention pattern on layers 13-16, scales $1 \times 1$, $2 \times 2$, $4 \times 4$.}
    \label{fig:cross_attn2}
\end{figure*}

We demonstrate comprehensive self- and cross-attention patterns across several layers and scales in Figures \ref{fig:self_attn}, \ref{fig:cross_attn1}, and \ref{fig:cross_attn2}. These visualizations suggest that NAMVIS learns correspondence patterns consistent with the underlying camera geometry.

\section{Evaluation Scenes}
\label{app:evaluation_scenes}

For reproducibility, we list the exact scenes used for evaluation on 
OmniObject3D (OO3D), GSO, and Objaverse. All methods are evaluated on the same 
scene list. Source and target camera configurations are fixed across methods 
whenever supported; for methods with camera-protocol constraints, we follow the 
protocol described in the main text.

\section{Broader Impacts}
Our work on NAMVIS presents both positive and negative potential societal impacts. On the positive side, by providing a highly efficient, diffusion-free alternative for multi-view generation, our method significantly lowers the computational barrier and energy footprint required for 3D content creation. This has the potential to democratize applications in augmented and virtual reality, education, and digital design. Conversely, the ability to rapidly generate realistic, geometrically consistent multi-view content from sparse inputs carries potential risks. These include the facilitation of deceptive 3D media, the unauthorized replication of copyrighted intellectual property or artist styles, and the potential generation of harmful digital assets. Addressing these negative implications will necessitate ongoing community efforts to develop robust 3D watermarking, provenance tracking, and responsible deployment guidelines.

\renewcommand{\arraystretch}{1.08}

\begin{longtable}{@{} S S S @{}}
\caption{Evaluation scenes from OmniObject3D (OO3D).}
\label{tab:eval_scenes_oo3d}\\
\toprule
\textbf{Scene} & \textbf{Scene} & \textbf{Scene} \\
\midrule
\endfirsthead
\toprule
\textbf{Scene} & \textbf{Scene} & \textbf{Scene} \\
\midrule
\endhead
\scene{chair_002} & \scene{chair_009} & \scene{chair_017} \\
\scene{chair_026} & \scene{backpack_003} & \scene{backpack_016} \\
\scene{backpack_022} & \scene{backpack_033} & \scene{helmet_004} \\
\scene{helmet_011} & \scene{helmet_013} & \scene{kettle_002} \\
\scene{kettle_008} & \scene{kettle_017} & \scene{toy_animals_007} \\
\scene{toy_animals_025} & \scene{toy_animals_049} & \scene{toy_animals_071} \\
\scene{apple_006} & \scene{apple_024} & \scene{apple_058} \\
\scene{vase_005} & \scene{vase_012} & \scene{vase_016} \\
\scene{bottle_001} & \scene{bottle_030} & \scene{bottle_032} \\
\scene{shoe_002} & \scene{shoe_003} & \scene{shoe_004} \\
\bottomrule
\end{longtable}

\begin{longtable}{@{} T T @{}}
\caption{Evaluation scenes from GSO.}
\label{tab:eval_scenes_gso}\\
\toprule
\textbf{Scene} & \textbf{Scene} \\
\midrule
\endfirsthead
\toprule
\textbf{Scene} & \textbf{Scene} \\
\midrule
\endhead
\scene{3D_Dollhouse_Swing} &
\scene{Netgear_Nighthawk_X6_AC3200_TriBand_Gigabit_Wireless_Router} \\

\scene{Hasbro_Dont_Wake_Daddy_Board_Game} &
\scene{BUILD_A_ZOO} \\

\scene{LEGO_5887_Dino_Defense_HQ} &
\scene{Melissa_Doug_Traffic_Signs_and_Vehicles} \\

\scene{New_Super_Mario_BrosWii_Wii_Game} &
\scene{ALPHABET_AZ_GRADIENT} \\

\scene{Crayola_Crayons_120_crayons} &
\scene{Tag_Dishtowel_Basket_Weave_Red_18_x_26} \\

\scene{VANS_FIRE_ROASTED_VEGGIE_CRACKERS_GLUTEN_FREE} &
\scene{Office_Depot_HP_920XL_920_High_Yield_Black_and_Standard_CMY_Color_Ink_Cartridges} \\

\scene{Android_Figure_Chrome} &
\scene{Black_Decker_Stainless_Steel_Toaster_4_Slice} \\

\scene{Metallic_Gold_Tieks_Italian_Leather_Ballet_Flats} &
\scene{Canon_225226_Ink_Cartridges_BlackColor_Cyan_Magenta_Yellow_6_count} \\

\scene{Seagate_1TB_Backup_Plus_portable_drive_Silver} &
\scene{Asus_Sabertooth_Z97_MARK_1_Motherboard_ATX_LGA1150_Socket} \\

\scene{Playmates_nickelodeon_teenage_mutant_ninja_turtles_shredder} &
\scene{Schleich_Bald_Eagle} \\

\scene{Philips_60ct_Warm_White_LED_Smooth_Mini_String_Lights} &
\scene{Shaxon_100_Molded_Category_6_RJ45RJ45_Shielded_Patch_Cord_White} \\

\scene{SpiderMan_Titan_Hero_12Inch_Action_Figure_5Hnn4mtkFsP} &
\scene{JBL_Charge_Speaker_portable_wireless_wired_Green} \\

\scene{Animal_Planet_Foam_2Headed_Dragon} &
\scene{Schleich_Spinosaurus_Action_Figure} \\

\scene{Down_To_Earth_Orchid_Pot_Ceramic_Lime} &
\scene{Tory_Burch_Kiernan_Riding_Boot} \\

\scene{Sootheze_Toasty_Orca} &
\scene{Mens_Authentic_Original_Boat_Shoe_in_Navy_Leather_RpT4GvUXRRP} \\
\bottomrule
\end{longtable}

\begin{longtable}{@{} S S S @{}}
\caption{Evaluation scenes from Objaverse.}
\label{tab:eval_scenes_objaverse}\\
\toprule
\textbf{Scene} & \textbf{Scene} & \textbf{Scene} \\
\midrule
\endfirsthead
\toprule
\textbf{Scene} & \textbf{Scene} & \textbf{Scene} \\
\midrule
\endhead
\scene{000074a334c541878360457c672b6c2e} &
\scene{001cfadfb9204424bccc45501ce6b90e} &
\scene{0023717f4f564cc99f4ded70db04f590} \\

\scene{003199cc6ff2410cb2d8e6f8a9cbb163} &
\scene{003ebdf86df345d39dc166563229fb85} &
\scene{0051724d6efa42de84aaf8467629160f} \\

\scene{0002c6eafa154e8bb08ebafb715a8d46} &
\scene{001d1b57e9df4273bede948b26429429} &
\scene{0023b3edbc114be188ca9d8f729dfaaf} \\

\scene{003219b0bec442d29725847969e4b6bf} &
\scene{003ed2834130466da3a51bb7fdd6bde5} &
\scene{005f5630a54b442291aeb0a5d487353b} \\

\scene{000b76f2b03e44e8ab44e1a1614be0f4} &
\scene{001fe8adb25a49d0b2650a5401dde019} &
\scene{0025c5e2333949feb1db259d4ff08dbe} \\

\scene{0032696f5871429fbd0549d9628f812c} &
\scene{0046f208ef8d4988ba7bb9d297f29ec7} &
\scene{00616f328a8b4b5a8e689f61e70758b6} \\

\scene{00124bcf3ca3463fbe05f28218cc0f5c} &
\scene{0022a3197f9646acbb9041eff2d1f55c} &
\scene{0025d57953fc4c8a80a44e59294e6841} \\

\scene{0033322379a24798a6875a5cb2de54f5} &
\scene{0048e8224b174b759771e39ff521ee2e} &
\scene{0061788e0741400c82289337a24af4f6} \\

\scene{00184eec45fe45ffa3826e9202fe7306} &
\scene{0022d2e01b014d328294b828e48defa1} &
\scene{00286954e2d54db8bc7832cc8682b6ff} \\

\scene{00380c3f5cf548c9846faf3c42dfd6db} &
\scene{004d02243a5b4117afc4baa45eb1eba0} &
\scene{0064add4992b426cb2f862e5875ebf6d} \\
\bottomrule
\end{longtable}

\clearpage

\end{document}